\documentclass[11pt]{article}

\usepackage{amsmath,amsfonts,amsthm,amssymb,amscd,mathrsfs,bm}
\usepackage{multirow}

\usepackage{graphicx} 
\usepackage{subfig} 

\usepackage{algorithm}
\usepackage{algorithmic}
\usepackage[left=1in, right=1in, top=1in, bottom=1in]{geometry}

\usepackage{hyperref}

\renewcommand{\algorithmicrequire}{\textbf{Input:}}
\renewcommand{\algorithmicensure}{\textbf{Output:}}

\graphicspath{{./pics/}}

\title{A Harmonic Extension Approach for Collaborative Ranking}
\author{
Da Kuang\thanks{Department of Mathematics, University of California, Los Angeles (UCLA), Los Angeles, CA, 90095. Email: \texttt{\{dakuang,sjo,bertozzi\}@math.ucla.edu}.}
\and
Zuoqiang Shi\thanks{Yau Mathematical Sciences Center, Tsinghua University, Beijing, China, 100084. Email: \texttt{zqshi@math.tsinghua.edu.cn}.}
\and
Stanley Osher$^*$
\and
Andrea Bertozzi$^*$
}
\date{}

\begin{document} 
\maketitle



\begin{abstract} 
We present a new perspective on graph-based methods for collaborative ranking for recommender systems. Unlike user-based or item-based methods that compute a weighted average of ratings given by the nearest neighbors, or low-rank approximation methods using convex optimization and the nuclear norm, we formulate matrix completion as a series of semi-supervised learning problems, and propagate the known ratings to the missing ones on the user-user or item-item graph globally. The semi-supervised learning problems are expressed as Laplace-Beltrami equations on a manifold, or namely, {\em harmonic extension}, and can be discretized by a point integral method. We show that our approach does not impose a low-rank Euclidean subspace on the data points, but instead minimizes the dimension of the underlying manifold. Our method, named {\em LDM (low dimensional manifold)}, turns out to be particularly effective in generating rankings of items, showing decent computational efficiency and robust ranking quality compared to state-of-the-art methods.
\end{abstract} 

\section{Introduction}
\label{intro}
Recommender systems are crucial components in contemporary e-commerce platforms (Amazon, eBay, Netflix, etc.), and were popularized by the Netflix challenge.
Detailed surveys of this field can be found in \cite{rssurvey1, rssurvey2}.
Recommendation algorithms are commonly based on collaborative filtering, or ``crowd of wisdom'', and can be categorized into memory-based and model-based approaches. Memory-based approaches include user-based and item-based recommendation \cite{itembased}. For example, for a user $u$, we retrieve the highly-rated items from the nearest neighbors of $u$, and recommend those items that have not been consumed by $u$. Memory-based methods are actually based on a graph, where a user-user or item-item similarity matrix defines the nearest neighbors of each user or item. In contrast, model-based methods are formulated as matrix completion problems which assume that the entire user-by-item rating matrix is low-rank \cite{svdcf}, and the goal is to predict the missing ratings given the observed ratings. While memory-based methods are typically more computationally efficient, model-based methods can achieve much higher quality for collaborative filtering.

Popular model-based methods such as regularized SVD \cite{svdcf} minimize the sum-of-squares error over all the observed ratings. When evaluating the predictive accuracy of these algorithms, we often divide the whole data set into a training set and a test set. After obtaining a model on the training set, we evaluate the accuracy of the model's prediction on the test set in order to see how well it generalizes to unseen data. However, the measure for evaluating success in a practical recommender system is very different. What we care more about is whether the top recommended items for a user $u$ will actually be ``liked'' by $u$. In an experimental setting, the evaluation measure for success in this context is the resemblance between the ranked list of top recommended items and the ranked list of observed ratings in the test set. Thus, this measure that compares two rankings is more relevant to the performance in real scenarios. The problem that places priority on the top recommended items rather than the absolute accuracy of predicted ratings is referred to as top-N recommendation \cite{eval_topn2}, or more recently {\em collaborative ranking} \cite{lcr, altsvm}, and is our focus in this paper.


We start with the matrix completion problem and formulate it as a series of semi-supervised learning problems, or in particular, 
harmonic extension problems on a manifold that can be solved by label propagation \cite{zgl, harmonic}. For each item, we want to know the ratings by all the users, and the goal of the semi-supervised 
learning problem is to propagate the known labels for this item (observed ratings) to the unknown labels on the user-user graph; 
and reversely, for each user, to propagate the known labels given by this user to the unknown labels on the item-item graph. 

Without loss of generality, we assume that there exists a user manifold,
denoted as $\mathcal{M}$, which consists of an infinite number of users. In a user-by-item rating system with $n$ items, each user
is identified by an $n$-dimensional vector that consists of the ratings to $n$ items. Thus, the user manifold $\mathcal{M}$ is a submanifold embedded in $\mathbb{R}^n$.
For the $i$-th item, we define the rating function $f_i: \mathcal{M}\rightarrow \mathbb{R}$ that maps a user into the rating of this item.

One basic observation is 
that for a fixed item, {\it similar users should give similar ratings}. This implies that the function $f_i, \; 1\le i\le n$ is a {\em smooth} function on $\mathcal{M}$. 
Therefore, it is natural to find the rating function $f_i$ by minimizing the following energy functional:
\begin{equation}
\label{energy}
E(f) = \int_{\mathcal{M}} \|\nabla_{\mathcal{M}} f(u)\|^2 d u,
\end{equation}
where $\nabla_{\mathcal{M}} f(u)$ is the gradient at $u$ defined on the manifold $\mathcal{M}$.
Using standard variational approaches, minimizing the above functional (\ref{energy}) is reduced to solving the Laplace-Beltrami equation on the user manifold $\mathcal{M}$. Then the Laplace-Beltrami
equation can be solved by a novel point integral method \cite{harmonic}.

For the harmonic extension model, we also have an interpretation based on the low dimensionality of the user manifold, after which we call our method {\em LDM}.
 The user manifold is a manifold embedded in $\mathbb{R}^n$, and $n$ is usually a large number. Compared to $n$, the intrinsic dimension of the user manifold is typically much smaller. Based on this observation, 
we use the dimension of the user manifold as a regularization to recover the rating matrix. This idea implies the following
 optimization problem:
\begin{align}
  \label{eq:model-manifold-intro}
 & \min_{X\in \mathbb{R}^{m\times n},\atop \mathcal{M}\subset \mathbb{R}^n} \quad\dim(\mathcal{M}),\\
 &\text{subject to:}\quad P_{\Omega}(X)=P_{\Omega}(A)  , \quad \mathcal{U}(X)\subset \mathcal{M},\nonumber
\end{align}
where $\dim(\mathcal{M})$ is the dimension of the manifold $\mathcal{M}$, and $\mathcal{U}(X)$ is the user set corresponding to the rows of $X$. 
$\Omega=\{(i,j) : \text{user }i\text{ rated item }j\}$ is the index set of the observed ratings, and $P_\Omega$ is the projection operator to $\Omega$,
\begin{align*}
  P_\Omega(X)=\left\{\begin{array}{cc}
x_{ij},& (i,j)\in \Omega,\\
0,& (i,j)\notin \Omega.
\end{array}\right.
\end{align*}
By referring to the theory in differential geometry, this optimization problem is reduced to the same formulation as that in harmonic extension (\ref{energy}) which gives our model a geometric 
interpretation.



Another important aspect of our proposed method is the weight matrix that defines the user-user or item-item graph. Because the information given in the rating matrix is incomplete, we can only assume that the weight matrix used in harmonic extension is a good guess. We will propose an efficient way to construct the weight matrix based on incomplete data.

Our main contribution is summarized as follows:
\begin{itemize}
\item We propose an algorithm that exploits manifold structures to solve the harmonic extension problem for collaborative ranking, representing a new perspective on graph-based methods for recommender systems.
\item On real data sets, our method achieves robust ranking quality with reasonable run-time, compared to state-of-the-art methods for large-scale recommender systems.
\end{itemize}

The rest of this paper is organized as follows. In Section \ref{harmonic}, we formulate the matrix completion problem as harmonic extension. In Section \ref{discretization}, we describe the point integral method to rigorously solve the discretized harmonic extension problem. In Section \ref{LDMM}, we show that our approach seeks to minimize the dimension of the underlying manifold. In Section \ref{sim_matrix}, we describe our more efficient way to compute the similarity matrix. In Section \ref{experiments}, we empirically demonstrate the efficiency and ranking quality of our method. In Section \ref{related}, we explain the connection and difference between our method and previous work. In Section \ref{conclusion}, we discuss our proposed method and its implication on other methods.

Here are some notations we will use. For a vector $x=[x_1,\cdots,x_m]^T$, we call $y=[x_{i_1},x_{i_2},\cdots,x_{i_r}]^T$ a {\em subvector} of length $r$ by extracting the elements of $x$ in the index set $\{i_1,\cdots,i_r\}$, where $i_1<i_2<\cdots<i_r$. For a matrix $M$, a vector $x$, integers $i,j$, and sets of row and column indices $S,S'$, we use $M_{i,j},M_{S,S'},M_{:,j},M_{S,j},x_S$ to denote an entry of $M$, a submatrix of $M$, the $j$-th column of $M$, a subvector of the $j$-th column of $M$, and a subvector of $x$, respectively.

\section{Harmonic Extension Formulation}
\label{harmonic}
Consider a user-by-item rating matrix $A=(a_{ij}) \in \mathbb{R}^{m \times n}$, where rows correspond to $m$ users, and columns correspond to $n$ items. 
The observed ratings are indexed by the set $\Omega=\{(i,j) : \text{user }i\text{ rated item }j\}$. Let $\Omega_i=\{1\le j\le m: (i,j)\in \Omega\}, \; 1\le i\le n$.
Suppose there exists a ``true'' rating matrix $A^*$ given by an oracle with no missing entries, which is not known to us, and $A|_{\Omega} = A^*|_{\Omega}$.

As mentioned in the introduction, we formulate matrix completion as a harmonic extension problem on a manifold. 
Recall the user manifold in Section \ref{intro},
denoted as $\mathcal{M}$, which is embedded in $\mathbb{R}^n$. 
The set of $m$ users in our user-by-item rating system is represented as $U=\{u_j, 1\le j\le m\}$ where $u_j$ is the $j$-th row of $A^*$
and $U\subset \mathcal{M}$ is a sample of $\mathcal{M}$. Let $U_i=\{u_j\in U: j\in \Omega_i\}$ be the collection of users who rate the $i$-th item. 

Then we compute the rating function $f_i$ for all the users by minimizing the energy functional in \eqref{energy}:
\begin{align}
\label{minenergy}
&\min_{f_i\in H^1(\mathcal{M})} \quad E(f_i) \quad\text{  subject to:  }\quad  f_i(u_j)|_{U_i}=a_{ij},
\end{align}
where $H^1$ is the Sobolev space.
Hence, we need to solve the following type of optimization problem for $n$ times. 
\begin{equation}
\label{opt}
\min_{f\in H^1(\mathcal{M})} \quad E(f) \quad \text{  subject to:  }\quad  f(u)|_{\Lambda}=g(u),
\end{equation}
where $\Lambda\subset\mathcal{M}$ is a point set.

To solve the above optimization problem, we first use the Bregman iteration to enforce the constraint \cite{bregman}.
\begin{itemize}
\item Solve
\begin{equation}
\label{opt-bregman} 
f^{k+1}=\arg\min_{f} \; E(f) +\mu  \|f-g+d^k\|_{L^2(\Lambda)}^2,
\end{equation}
where $\|f\|_{L^2(\Lambda)}^2=\sum_{u\in \Lambda}|f(u)|^2$, $d^n$ is a function defined on $\Lambda$.
\item Update $d^k$,
$$d^{k+1}(u)=d^k(u) + f^{k+1}(u)-g(u),\; \forall u\in \Lambda.$$
\item Repeat above process until convergence.
\end{itemize}

\newcommand{\bx}{\mathbf{x}}
\newcommand{\bb}{\mathbf{b}}
\newcommand{\bn}{\mathbf{n}}
\newcommand{\by}{\mathbf{y}}

Using a standard variational approach, the solution to (\ref{opt-bregman}) can be reduced to the following Laplace-Beltrami equation:
\begin{equation}
\label{laplace}
\begin{cases}
\displaystyle \Delta_{\mathcal{M}} f(\bx)-\mu \sum_{\by\in \Lambda} \delta(\bx-\by)(f(\by)-h(\by)) = 0, \; \bx \in \mathcal{M}, \\
\displaystyle\frac{\partial f}{\partial \mathbf{n}}(\bx)=0, \quad \bx \in \partial \mathcal{M},
\end{cases}
\end{equation}
where $\delta$ is the Dirac-$\delta$ function in $\mathcal{M}$, $h=g-d^n$ is a given function on $\Lambda$, and $\mathbf{n}$ is the outer normal vector.
That is to say, the function $f$ that minimizes (\ref{opt}) is a harmonic function on $\mathcal{M} \backslash \partial \mathcal{M}$, and (\ref{laplace}) is called the {\em harmonic extension} problem in the continuous setting.



If the underlying manifold $\mathcal{M}$ (the true rating matrix $A^*$ in the discrete setting) were known, the $n$ problems in (\ref{minenergy}) would be independent with each other and could be solved individually by (\ref{laplace}). However, $\mathcal{M}$ is not known, and therefore we have to get a good estimate for the operator $\Delta_{\mathcal{M}}$ based on $f_j$'s. Our algorithm for solving (\ref{minenergy}) is described on a high level in Algorithm \ref{alg:continuous_case}, where we iteratively update $f_j$'s and our estimate for $\Delta_{\mathcal{M}}$.

In the next section, we use the point integral method (PIM) to solve the Laplace-Beltrami equation \eqref{laplace}.

\begin{algorithm}[!t]
\caption{Algorithm for solving (\ref{minenergy})}
\label{alg:continuous_case} \centering
\begin{algorithmic}[1]
\REPEAT
\STATE Get estimate for $\Delta_{\mathcal{M}}$ based on $f_1,f_2,\cdots,f_n$
\FOR {$i=1$ to $n$}
\STATE Solve (\ref{laplace}) to obtain $f_i$
\ENDFOR
\UNTIL {some stopping criterion is satisfied}
\end{algorithmic}
\end{algorithm}

\textbf{Remark.} The update of $f_j$'s in Algorithm \ref{alg:continuous_case} follows the ``Jacobi'' scheme. We could also use the ``Gauss-Seidel'' scheme, i.e. re-estimate $\Delta_{\mathcal{M}}$ after the update of each $f_j$, but that would be much slower.


\section{Point Integral Method (PIM)}
\label{discretization}
\newcommand{\p}{\partial}
\newcommand{\B}{\mathbf}
\newcommand{\M}{\mathcal{M}}

\newcommand{\mathd}{\mathrm{d}}

\newcommand{\bw}{\mathbf{w}}
\newcommand{\bfp}{\mathbf{p}}
\newcommand{\bp}{\mathbf{\Phi}}
\newcommand{\bs}{\mathbf{s}}
\newcommand{\al}{\alpha}
\newcommand{\e}{\epsilon}
\newcommand{\Bw}{\mbox{\boldmath$\omega$}}
\newcommand{\hr}{R\left(\frac{|\bx-\by|^2}{4t}\right)}
\newcommand{\hbr}{\bar{R}\left(\frac{|\bx-\by|^2}{4t}\right)}
\newcommand{\nhk}{\frac{1}{(4\pi t)^{k/2}}}
\newcommand{\tnhk}{\frac{1}{(4\pi t)^{k/2}t}}

\newcommand{\bV}{\mathbf{V}}
\newcommand{\bA}{\mathbf{A}}
\newcommand{\bu}{\mathbf{u}}
\newcommand{\bg}{\mathbf{g}}
\newcommand{\bh}{\mathbf{h}}

\newcommand{\hk}{R_t(\bx, \by)}
\newcommand{\rhk}{\bar{R}_t(\bx, \by)}
\newcommand{\hkpipj}{R_t({\bf p}_i, {\bf p}_j)}
\newcommand{\hkpisj}{R_t({\bf p}_i, {\bf s}_j)}
\newcommand{\rhkpipj}{\bar{R}_t({\bf p}_i, {\bf p}_j)}
\newcommand{\rhkpisj}{\bar{R}_t({\bf p}_i, {\bf s}_j)}
\newcommand{\hkxpj}{R_t(\bx, {\bf p}_j)}
\newcommand{\hkxsj}{R_t(\bx, {\bf s}_j)}
\newcommand{\rhkxpj}{\bar{R}_t(\bx, {\bf p}_j)}
\newcommand{\rhkxsj}{\bar{R}_t(\bx, {\bf s}_j)}
\newcommand{\bff}{{\bf f}}
\newcommand{\bfg}{{\bf g}}
\newcommand{\bfu}{{\bf u}}
\newcommand{\bfs}{{\bf s}}
\newcommand{\bfn}{{\bf n}}
\newcommand{\bz}{\mathbf{z}}
\newcommand{\bR}{\bar{R}}
\newcommand{\invt}{\frac{1}{t}}
\newcommand{\LP}{\mathcal{L}}
\newcommand{\W}{\mathcal{W}}
\newcommand{\D}{\mathcal{D}}

\newtheorem{theorem}{\textbf{Theorem}}[section]
\newtheorem{lemma}{\textbf{Lemma}}[section]
\newtheorem{remark}{\textbf{Remark}}[section]
\newtheorem{proposition}{\textbf{Proposition}}[section]
\newtheorem{assumption}{\textbf{Assumption}}[section]
\newtheorem{definition}{\textbf{Definition}}[section]
\newtheorem{corollary}{\textbf{Corollary}}[section]

\paragraph{Integral Equation:~}
The key observation in PIM is that the Laplace-Beltrami operator has the following integral approximation:
\begin{align}
&\int_{\M}w_t(\bx,\by)\Delta_{\M}f(\by)\mathd \by\nonumber
\\\approx &
-\frac{1}{t}\int_{\M} (f(\bx)-f(\by))w_t(\bx, \by) \mathd \by\nonumber\\
&+2\int_{\p \M} \frac{\p f(\by)}{\p \bn}w_t(\bx, \by) \mathd \tau_\by,
\label{eqn:integral}
\end{align}
where $w_t(\bx, \by) = \exp(-\frac{|\bx-\by|^2}{4t})$. The following theorem gives the accuracy of the integral approximation.
\begin{theorem}
  \label{thm:local-error} \cite{neumann}
If $f\in C^3(\M)$ is a smooth function on $\M$,
then for any $\bx \in \M$,
\begin{eqnarray}
\label{eqn:local-error}
\left\|r(f) \right\|_{L^2(\mathcal{M})}=O(t^{1/4}),
\end{eqnarray}
where
\begin{align*}
  r(f)=&\int_{\M}w_t(\bx,\by)\Delta_{\M}f(\by)\mathd \by\\
&+\frac{1}{t}\int_{\M} (f(\bx)-f(\by))w_t(\bx, \by) \mathd \by \\
&- 2\int_{\p \M} \frac{\p f(\by)}{\p \bn} w_t(\bx, \by) \mathd \tau_\by.
\end{align*}
\end{theorem}
Applying the integral approximation \eqref{eqn:integral} to the Laplace-Beltrami equation \eqref{laplace}, we get an 
integral equation
\begin{align}
\frac{1}{t}\int_{\M} (f(\bx)&-f(\by))w_t(\bx, \by) \mathd \by\nonumber\\
&+\mu\sum_{\by\in \Lambda}w_t(\bx,\by)(f(\by)-h(\by))= 0,
\label{eqn:integral2}
\end{align}
In this integral equation, there are no derivatives, and therefore it is easy to discretize over the point cloud. 

\paragraph{Discretization:~}
We notice that the closed form of the user manifold $\M$ is not known, and we only have a sample of $\M$, i.e. $U$.
Next, we discretize the integral equation \eqref{eqn:integral} over the point set $U$. 

Assume that the point set $U$ is uniformly distributed over $\M$. The integral equation can be 
discretized easily, as follows: 
\begin{align}
  \label{eq:discretize}
  \frac{|\M|}{m}\sum_{j=1}^m &w_t(\bx_i,\bx_j)(f(\bx_i)-f(\bx_j))+\nonumber\\
&\mu t \sum_{\by\in \Lambda}w_t(\bx_i,\by)(f(\by)-h(\by))=0
\end{align}
where $|\M|$ is the volume of the manifold $\M$.

We can rewrite \eqref{eq:discretize} in the matrix form.
\begin{align}
  \label{eq:dis-matrix}
  \bm{L}\bm{f}+\bar{\mu} \bm{W}_{:,\Lambda}\bm{f}_\Lambda=\bar{\mu} \bm{W}_{:,\Lambda}\bm{h}.
\end{align}
where $\bm{h}=(h_1,\cdots,h_m)$ and $\bar{\mu}=\frac{\mu t m}{|\M|}$. $\bm{L}$ is a $m\times m$ matrix which is given as
\begin{align}
  \label{eq:matrix-L}
  \bm{L}=\bm{D}-\bm{W}
\end{align}
 where $\bm{W}=(w_{ij}),\; i,j=1,\cdots, m$ is the weight matrix and 
$\bm{D} = \mbox{diag}(d_i)$ with $d_i=\sum_{j=1}^m w_{ij}$.

\begin{algorithm}[tb]
\floatname{algorithm}{Algorithm}
\caption{Harmonic Extension}
\label{alg:harmonic}
\begin{algorithmic}[1]
\REQUIRE Initial rating matrix $A$.
\ENSURE  Rating matrix $R$.
\STATE Set $R=A$.
\REPEAT
\STATE Estimate the weight matrix $\bm{W}=(w_{ij})$ from the user set $U$ (Algorithm \ref{alg:sim_matrix}).
\STATE Compute the graph Laplacian matrix: $\bm{L}=\bm{D}-\bm{W}$
\FOR{$i=1$ to $n$}
\REPEAT
\STATE Solve the following linear systems
\begin{align*}
  \bm{L}\bm{f}_i+\bar{\mu} \bm{W}_{:,U_i}(\bm{f}_i)_{U_i}=\bar{\mu} \bm{W}_{:,U_i}\bm{h}_{U_i},
\end{align*}
where $\bm{h}=\bm{g}_i-\bm{d}^k$.
\STATE Update $\bm{d}^k$,
$$\bm{d}^{k+1}=\bm{d}^k+\bm{f}^{k+1}-\bm{g}_i$$
\UNTIL {some stopping criterion for the Bregman iterations is satisfied} 
\ENDFOR
\STATE $r_{ij}=f_i(u_j)$ and $R=(r_{ij})$.
\UNTIL {some stopping criterion is satisfied}
\end{algorithmic}
\end{algorithm}

\textbf{Remark.} In the harmonic extension approach, we use a continuous formulation based on the underlying user manifold. And the point integral method is used to 
solve the Laplace-Beltrami equation on the manifold. If a graph model were used at the beginning, the natural choice for harmonic extension would be the graph Laplacian. 
However, it has been observed that the graph Laplacian is not consisitent in solving the harmonic extension problem \cite{harmonic,LDMM_image}, and PIM gives much better results. 

\textbf{Remark.} The optimization problem we defined in (\ref{energy}) can be viewed as a continuous analog of the discrete harmonic extension problem \cite{zgl}, which we write in our notations:
\begin{equation}
\label{energy_discrete}
\min_{\bm{f}_i} \sum_{j,j'=1}^n w_{jj'} \left( (\bm{f}_i)_j - (\bm{f}_i)_{j'} \right)^2 \ \text{  subject to:  } \  (\bm{f}_i)_{U_i}=A_{U_i,i}.
\end{equation}
The formulation (\ref{energy_discrete}) in the context of collaborative ranking can be seen as minimizing the weighted sum of squared error in pairwise ranking. This form of loss function considers all the possible pairwise rankings of items, which is different from the loss function in previous work on collaborative ranking \cite{lcr,altsvm}:
\begin{equation}
\label{pairwiseloss}
\sum_{j,j' \in U_i} \mathcal{L}\left( [a_{ji}-a_{j'i}] - [(\bm{f}_i)_j - (\bm{f}_i)_{j'}] \right),
\end{equation}
where $\mathcal{L}$ is a loss function such as hinge loss and exponential loss. Only the pairwise rankings of items in the {\em training set} are considered in (\ref{pairwiseloss}).



\section{Low Dimensional Manifold (LDM) Interpretation}
\label{LDMM}
In this section, we emphasize the other interpretation of our method based on the low dimensionality of the user manifold. In the user-by-item rating system, a user is 
represented by an $n$-dimensional vector that consists of the ratings to $n$ items, and the user manifold is a manifold embedded in $\mathbb{R}^n$.
Usually, $n$, the number of items, is a large number in the order of $10^3\sim 10^6$. The intrinsic dimension of the user manifold is much less than $n$. 
Based on this observation, 
it is natural to recover the rating matrix by looking for the user manifold with the lowest dimension, which implies the optimization problem in (\ref{eq:model-manifold-intro}):
\begin{align*}
 & \min_{X\in \mathbb{R}^{m\times n},\atop \M\subset \mathbb{R}^n} \quad\dim(\mathcal{M}),\\
 &\text{subject to:}\quad P_{\Omega}(X)=P_{\Omega}(A)  , \quad \mathcal{U}(X)\subset \M.\nonumber
\end{align*}
where $\dim(\M)$ is the dimension of the manifold $\M$, and $\mathcal{U}(X)$ is the user set corresponding to the rows of $X$.

Next, we need to give a mathematical expression to compute $\dim(\mathcal{M})$.
Here we assume $\M$ is a smooth manifold embedded in $\mathbb{R}^n$.
Let $\alpha_i, \; i=1,\cdots,d$ be the coordinate functions on $\M$, i.e.
\begin{align}
  \label{eq:fun-coordinate}
  \alpha_i(\bx)=x_i,\quad \forall \bx=(x_1,\cdots,x_n)\in \M
\end{align}
Using differential geometry, we have the following formula \cite{LDMM_image}.
\begin{proposition} 
\label{prop:dim}
Let $\M$ be a smooth submanifold isometrically embedded in $\mathbb{R}^n$. For any $\bx\in \M$,  
  \begin{align*}
    \dim(\M)=\sum_{i=1}^n\|\nabla_\M\alpha_i(\bx)\|^2
  \end{align*}
where $\nabla_\M$ is the gradient in the manifold $\M$.
\end{proposition}

We can clearly see that $\alpha_i$ corresponds to the rating function $f_i$. Using the above proposition, the manifold dimension minimization problem \eqref{eq:model-manifold-intro} can be rewritten as
\begin{align}
  \label{eq:model-manifold-grad}
&  \min_{X\in \mathbb{R}^{m\times n},\atop \M\subset \mathbb{R}^d} \quad \sum_{i=1}^d\|\nabla_{\M} f_i\|_{L^2(\M)}^2,\\
& \text{subject to:}\quad
f_i(x_j)|_{U_i}=a_{ij}, \quad \mathcal{U}(X)\subset \M,\nonumber
\end{align}
where
\begin{equation}
  \label{eq:def-l2}
  \|\nabla_{\M} f_i\|_{L^2(\M)}=\left(\int_{\M} \|\nabla_{\M} f_i(\bx)\|^2\mathd {\bx}\right)^{1/2}.
\end{equation}
This is the same optimization problem we solved in Section \ref{harmonic} and Section \ref{discretization}.


\section{Weight Matrix}
\label{sim_matrix}
The weight matrix $\bm{W}$ plays an important role in our algorithm as well as other graph-based approaches \cite{zgl, itembased, llorma, mcgraph2014}. We employ the typical user-user or item-item graph with cosine similarity used in existing memory-based approaches for recommendation \cite{itembased}. However, we have made substantial changes to make the procedure efficient for large sparse rating matrices. Our algorithm for building the weight matrix is described in detail in Algorithm \ref{alg:sim_matrix}. Again, we consider the user-user graph without loss of generality.

First, as usual, we can only afford to compute and store a sparse nearest-neighbor weight matrix. To get the $K$ nearest neighbors for a target user $u$, traditional algorithms in memory-based methods require computing the distances between $u$ and every other user, and selecting the $K$ closest ones, where most of the computation is wasted if $K << m$. In our algorithm, we first identify the nearest neighbors approximately, without computing the actual distances or similarities, and then compute the similarities between $u$ and its nearest neighbors only. We use a binary rating matrix $R_B$ that records ``rated or not-rated'' information (Algorithm \ref{alg:sim_matrix}, line 1-2), and determine the $K$ nearest neighbors using an {\em kd}-tree based approximate nearest neighbor algorithm (line 3, line 5) \cite{kdtree}. That is to say, two users who have rated similar sets of movies are more likely to be considered to be in each other's neighborhood, regardless of their numeric ratings for those movies. Neither of the ways to build the {\em kd}-tree and to find nearest neighbors based on the tree are as precise as a na\"{i}ve search; however, empirical results in the next section have shown that our approximate strategy does not compromise the quality.

Second, we extended the VLFeat package \cite{vlfeat} to enable building a {\em kd}-tree from a sparse data matrix (in our case, $R_B$) and querying the tree with sparse vectors. {\em kd}-tree uses a space partitioning scheme for efficient neighbor search \cite{kdtree_orig}. For high-dimensional data, we employ the greedy way that chooses the most varying dimension for space partitioning at each step of building the tree \cite{kdtree}, and the procedure terminates when each leaf partition has one data point. Thus, the complexity of building the tree is not exponential, contrary to common understanding; and the practical performance of {\em kd}-tree can be very efficient and better than that of locality sensitive hashing \cite{lsh, rightknn}. For example, in our case with $m$ data points in $n$ dimensions, the complexity of building the tree is $O(m)$, rather than $O(2^n)$. In addition, when querying the tree, we put an upper bound $D$ on the maximum number of distance comparisons, and therefore the overall complexity of finding $K$ nearest neighbors for all the $m$ data points is $O(Dm\log K)$ (the $\log K$ factor comes from maintaining a heap data structure for the $K$ nearest neighbors).

Note that the resulting graph is not symmetric. Also the cosine similarity for two data points with incomplete information is defined by using the co-rated items only (Algorithm \ref{alg:sim_matrix}, line 8-9) \cite{itembased}.

\begin{algorithm}[!t]
\caption{Building weight matrix from incomplete rating data}
\label{alg:sim_matrix} \centering
\begin{algorithmic}[1]
\REQUIRE Incomplete rating matrix $R \in \mathbb{R}^{m \times n}$, number of nearest neighbors $K$
\STATE Generate binary rating matrix $R_B \in \mathbb{R}^{m \times n}$:
\begin{equation*}
(R_B)_{j,j'} = 
\begin{cases}
1, \quad R_{j,j'}\text{ is not missing} \\
0, \quad R_{j,j'}\text{ is missing}
\end{cases}
\end{equation*}
\STATE Normalize each row of $R_B$ such that $\|(R_B)_{j,:}\|_2=1$, $\forall j, 1 \leq j \leq m$
\STATE Build a {\em kd}-tree on the data points (rows) in $R_B$
\STATE Initialize a sparse matrix $\bm{W} \leftarrow \mathbf{0}^{m \times m}$
\FOR {$j=1$ to $m$}
\STATE $\mathcal{N}_B \leftarrow$ The set of $K$ approximate nearest neighbors of $(R_B)_{j,:}$, found by querying the {\em kd}-tree
\FOR {$j' \in \mathcal{N}_B$ ($j' \neq j$)}
\STATE Set of co-rated items $\mathcal{C} \leftarrow$ \\ $\{i: R_{j,i} \text{ is not missing, and } R_{j',i} \text{ is not missing}\}$
\STATE $\bm{W}_{j,j'} \leftarrow \text{cosine}(R_{j,\mathcal{C}}, R_{j',\mathcal{C}})$
\ENDFOR
\ENDFOR
\ENSURE Sparse weight matrix $\bm{W} \in \mathbb{R}^{m \times m}$
\end{algorithmic}
\end{algorithm}

\section{Experiments}
\label{experiments}
In this section, we evaluate our proposed method LDM in terms of both run-time and ranking quality. Since the volume of literature on recommender systems, collaborative filtering, and matrix completion is huge, we select only a few existing methods to compare with. All the experiments are run on a Linux laptop with one Intel i7-5600U CPU (4 physical threads) and 8 GB memory.

\subsection{Data Sets}

\begin{table}[!t]
\caption{Statistics of the data sets in our experiments. $N$ is the number of ratings in the training set for each user. The total number of ratings in the training set is $N\times$(\# users).}
\label{table:datasets} \centering
{\small
\begin{tabular}{|l|c|r|r|r|}
\hline
\multicolumn{2}{|c|}{Data set} & \# users & \# items & \# ratings \\
\hline
\multirow{2}{*}{MovieLens-100k} & $N=10$ & 943 & 1,682 & 100,000 \\
 & $N=20$ & 744 & 1,682 & 95,269 \\
\hline
\multirow{3}{*}{MovieLens-1m} & $N=10$ & 6,040 & 3,706 & 1,000,209 \\
 & $N=20$ & 5,289 & 3,701 & 982,040 \\
 & $N=50$ & 3,938 & 3,677 & 924,053 \\
\hline
\multirow{4}{*}{MovieLens-10m} & $N=10$ & 69,878 & 10,677 & 10,000,054 \\
& $N=20$ & 57,534 & 10,675 & 9,704,223 \\
& $N=50$ & 38,604 & 10,672 & 8,897,688 \\
 & $N=100$ & 24,328 & 10,666 & 7,730,011 \\
\hline
\end{tabular}}
\end{table}

We use three MovieLens\footnote{\url{http://grouplens.org/datasets/movielens/}} data sets in our experiments: {\em MovieLens-100k}, {\em MovieLens-1m}, and {\em MovieLens-10m}. In each of the data sets, each user has at least 20 ratings. Following the convention in previous work on collaborative ranking, we randomly select $N$ ratings for each user as the training set, and the other ratings were used for testing. To keep at least 10 ratings in the testing set for each user, we remove the users with fewer than $N+10$ ratings. After this preprocessing, we can generate several versions of these data sets with different $N$'s, whose information is summarized in Table \ref{table:datasets}.

\subsection{Methods for Comparison}
\label{algcompare}

We compare our LDM method with singular value decomposition (SVD) as a baseline, and two state-of-the-art methods that are designed specifically for collaborative ranking. All the three methods for comparison optimize a pairwise ranking loss function. We do not perform hyperparameter selection on a separate validation set because it is time-consuming, but we investigate the effect of the hyperparameters in Section \ref{hyperparameters}; and in Section \ref{results}, we use a fixed set of hyperparameters that can achieve a good balance between run-time and ranking quality, which are found empirically across several data sets. We list these methods below (their program options that will be used in Section \ref{results} is in the footnote):
\begin{itemize}
\item {\em SVD}: We use the Java implementation of ranking-based SVD in the PREA toolkit\footnote{\url{http://prea.gatech.edu}}$^,$\footnote{The command-line options for SVD is \texttt{-a ranksvd exp\_add 5 5 25}.}. This version uses gradient descent to optimize the ranking-based loss function (\ref{pairwiseloss}), as opposed to the squared loss function in regularized SVD.
\item {\em LCR}: Local collaborative ranking \cite{lcr}, as implemented in the PREA toolkit\footnote{The command-line options for LCR is \texttt{-a pgllorma exp\_add 5 5 13}.}.
\item {\em AltSVM}: Alternating support vector machine \cite{altsvm}, as implemented in the \texttt{collranking} package\footnote{\url{https://github.com/dhpark22/collranking}}. We employ the default configurations.
\end{itemize}

Lastly, our proposed method LDM (Algorithm \ref{alg:harmonic}) is implemented in Matlab, and the construction and querying of the {\em kd}-tree (Algorithm \ref{alg:sim_matrix}) is implemented in C, for which we extended the VLFeat\footnote{\url{http://www.vlfeat.org/}} package to build a {\em kd}-tree with a sparse input matrix efficiently. In Algorithm \ref{alg:harmonic}, we set $\mu=1$ and run one inner iteration and one outer iteration only, since we empirically found that the weight matrix constructed from the incomplete input ratings by Algorithm \ref{alg:sim_matrix} is often good enough for the algorithm to converge in one iteration. Algorithm \ref{alg:harmonic} typically accounts for most ($\sim95$\%) of the run-time in our method.

All the programs except SVD use 4 threads in our experiments, and are applied to the same training and testing random splits.

\subsection{Evaluation Measure}
\label{sec:eval_measure}

We evaluate the ranking quality by {\em normalized discounted cumulative gain} (NDCG) @$K$ \cite{ndcg}, averaged over all the users. Given a ranked list of $t$ items $i_1,\cdots,i_t$ and their {\em ground-truth} relevance score $r_{i_1},\cdots,r_{i_t}$, the DCG@$K$ score ($K \leq t$) is computed as
\begin{equation}
\label{dcg}
\text{DCG@}K (i_1,\cdots,i_t) = \sum_{j=1}^K {2^{r_{i_j}} - 1 \over \log_2(j+1)}.
\end{equation}
Then, we sort the list of items in the decreasing order of the relevance score and obtain the list $i_1^*,\cdots,i_t^*$, where $r_{i_1^*} \geq \cdots \geq r_{i_t^*}$, and this sorted list achieves the maximum DCG@$K$ score over all the possible permutations. The NDCG score is defined as the normalized version of (\ref{dcg}):
\begin{equation}
\label{ndcg}
\text{NDCG@}K (i_1,\cdots,i_t) = {\text{DCG@}K (i_1,\cdots,i_t) \over \text{DCG@}K (i_1^*,\cdots,i_t^*)}.
\end{equation}
We only evaluate NDCG@$10$ due to limited space. Note that for a user $u$, the list of items that is provided to compute (\ref{ndcg}) is {\em all} the items with observed ratings given by $u$ in the test set, not only the highest-ranking ones.

In contrast to previous work on top-N recommender systems \cite{eval_topn,eval_topn2}, we discourage the use of Precision@$K$ in the context of collaborative ranking for recommender systems, which measures the proportion of actually rated items out of the top $K$ items in the ranked list of all the items in the data set. Contemporary recommender systems typically use numeric ratings rather than binary ratings. In a 1-to-5 rating system, for example, a 1-star item should be less favorable than an unrated item with an expected 3-star rating. However, a recommender system that ranks 1-star items at the top positions would get a higher Precision@$K$ score than one that ranks unrated items at the top positions. Thus, Precision@$K$ is not a valid measure for collaborative ranking with a non-binary rating system.


\subsection{Effect of Parameter Selection}
\label{hyperparameters}

\begin{figure*}[!t]
\includegraphics[width=0.33\textwidth]{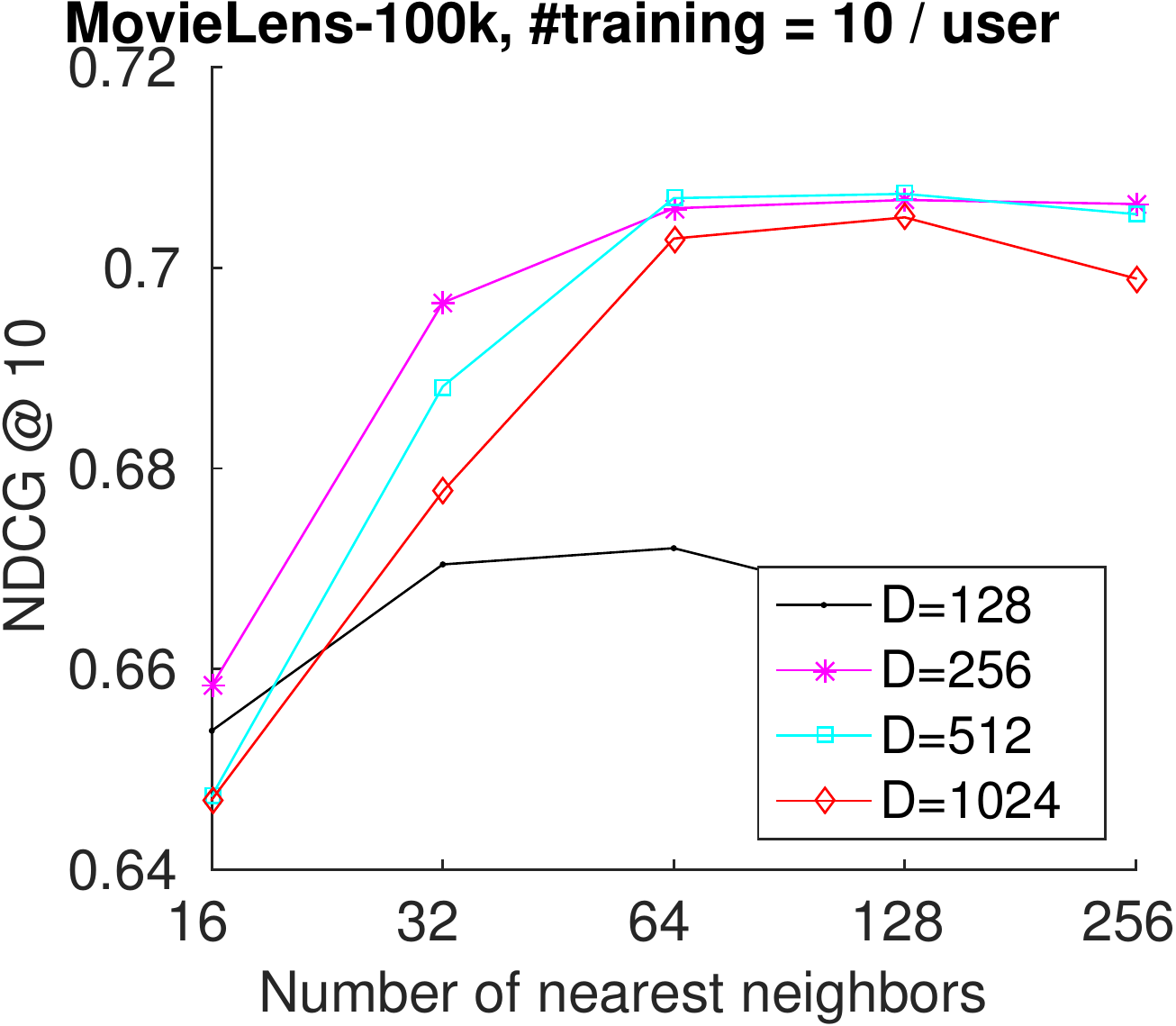}
\includegraphics[width=0.33\textwidth]{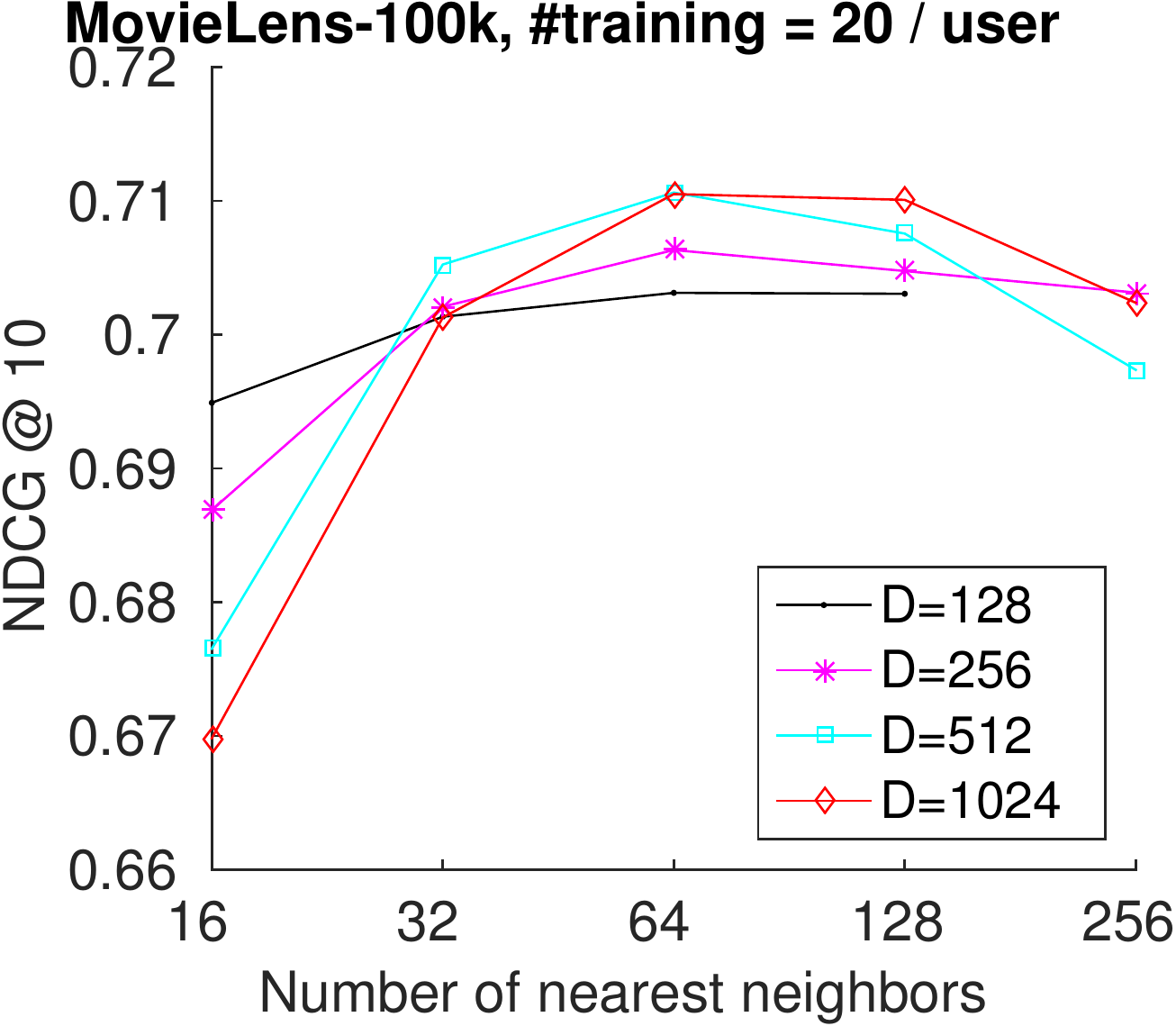}
\includegraphics[width=0.33\textwidth]{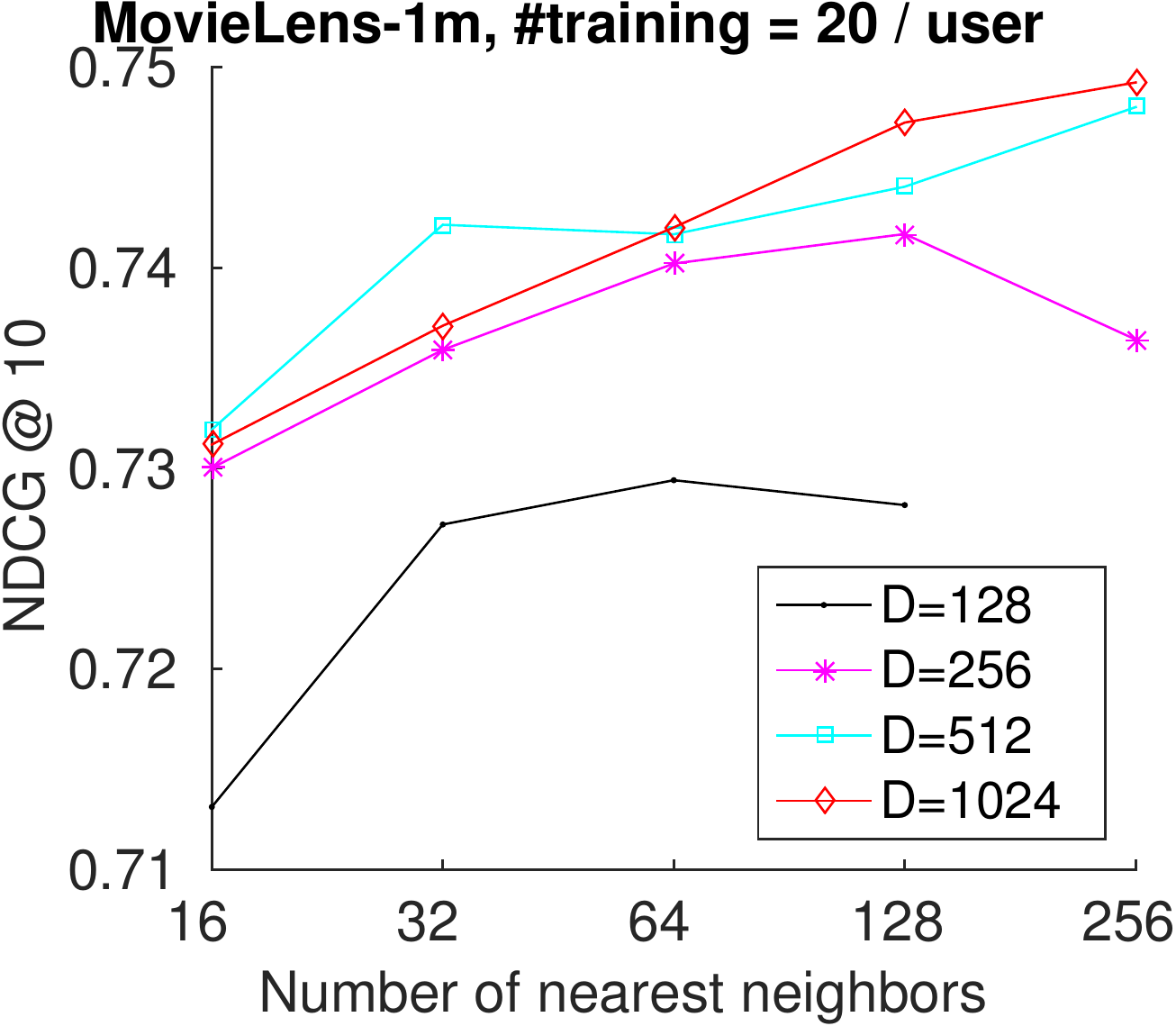}
\caption{Ranking quality NDCG@$10$ (Section \ref{sec:eval_measure}) as a function of the number of nearest neighbors ($k$) when constructing the {\em kd}-tree and the maximum number of comparisons ($D$) when querying the {\em kd}-tree. $k$ must not be larger than $D$.}
\label{fig:kdtree_param} \centering
\end{figure*}

\begin{figure*}[!t]
\includegraphics[width=0.32\textwidth]{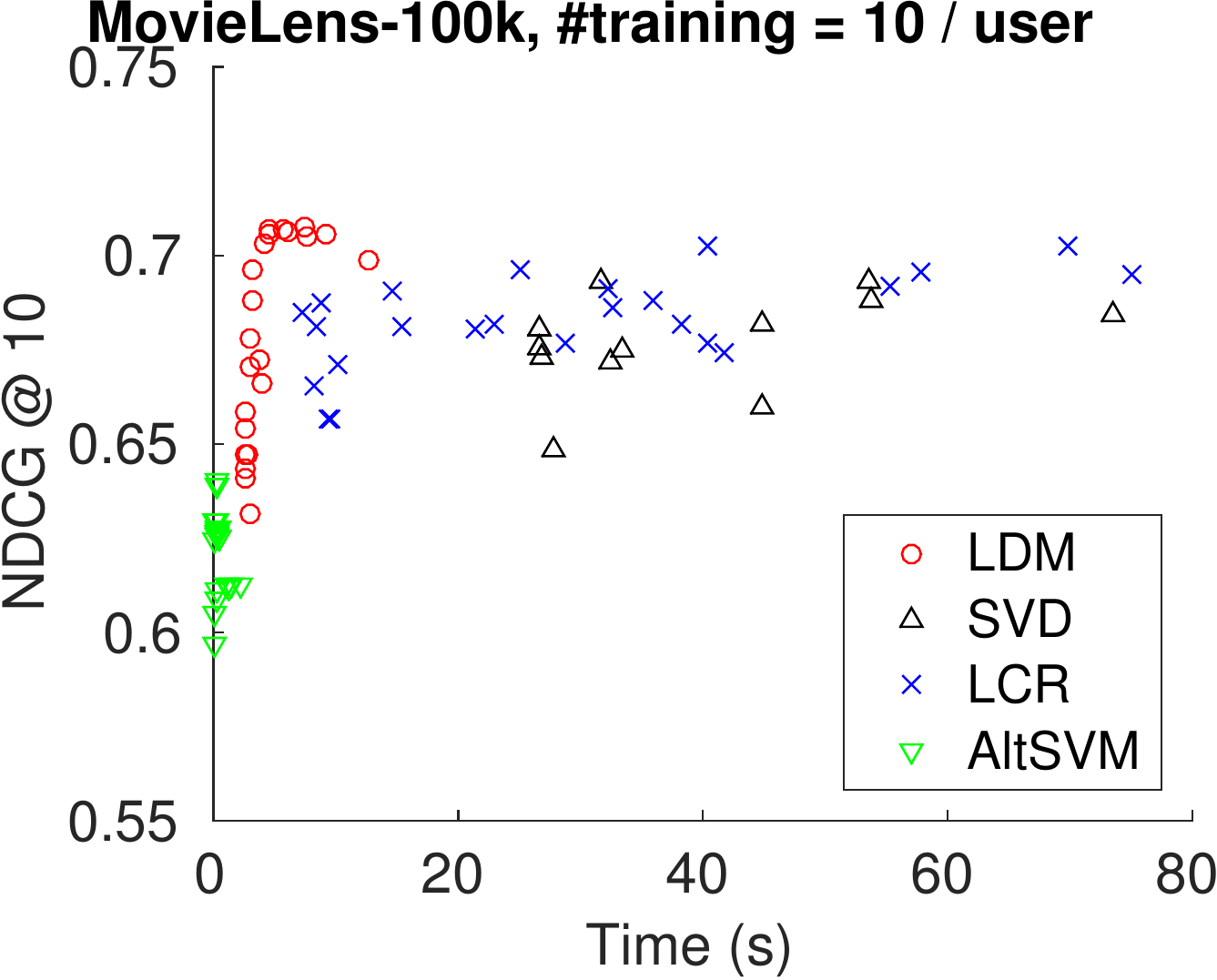}
\includegraphics[width=0.34\textwidth]{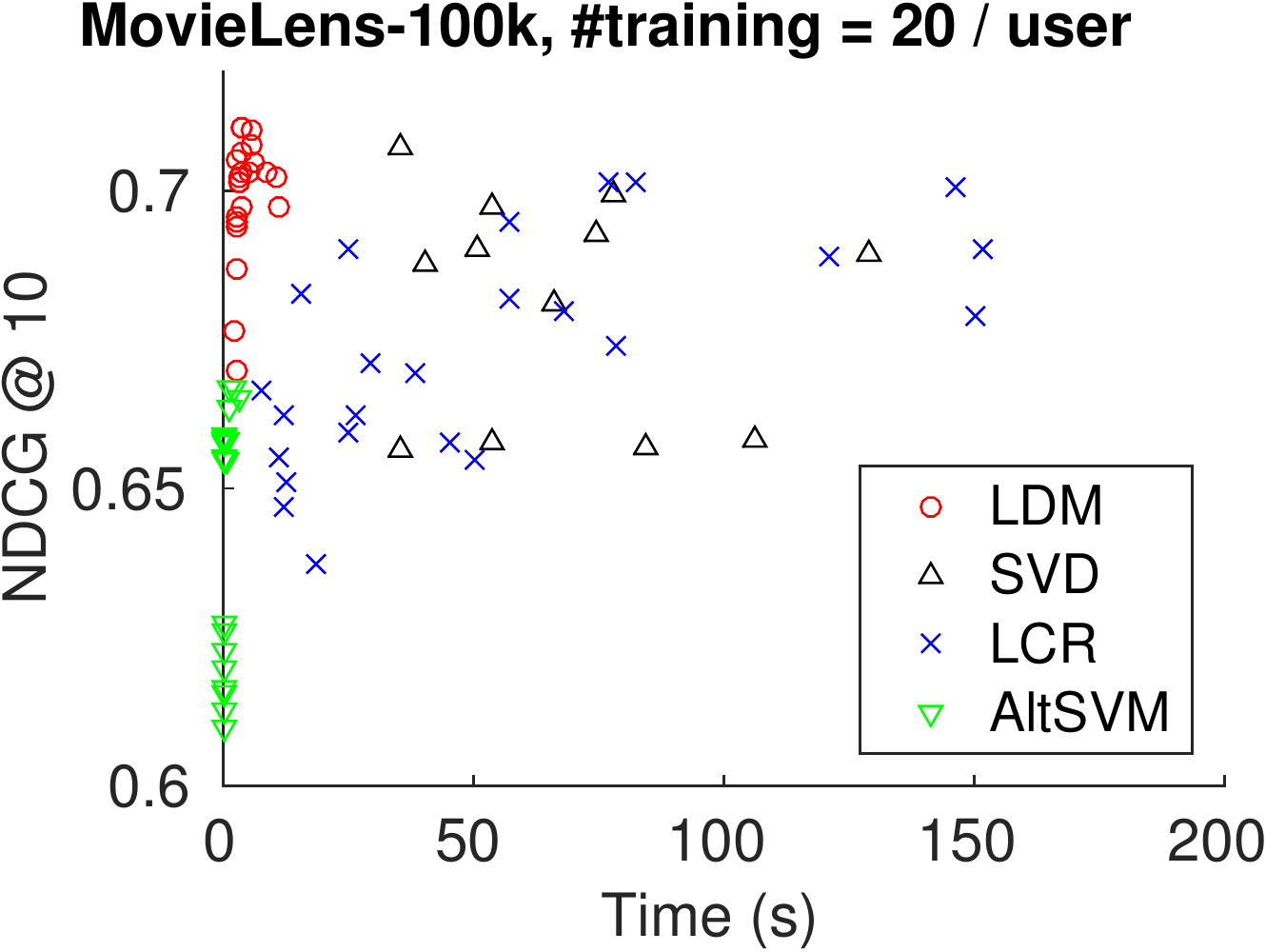}
\includegraphics[width=0.325\textwidth]{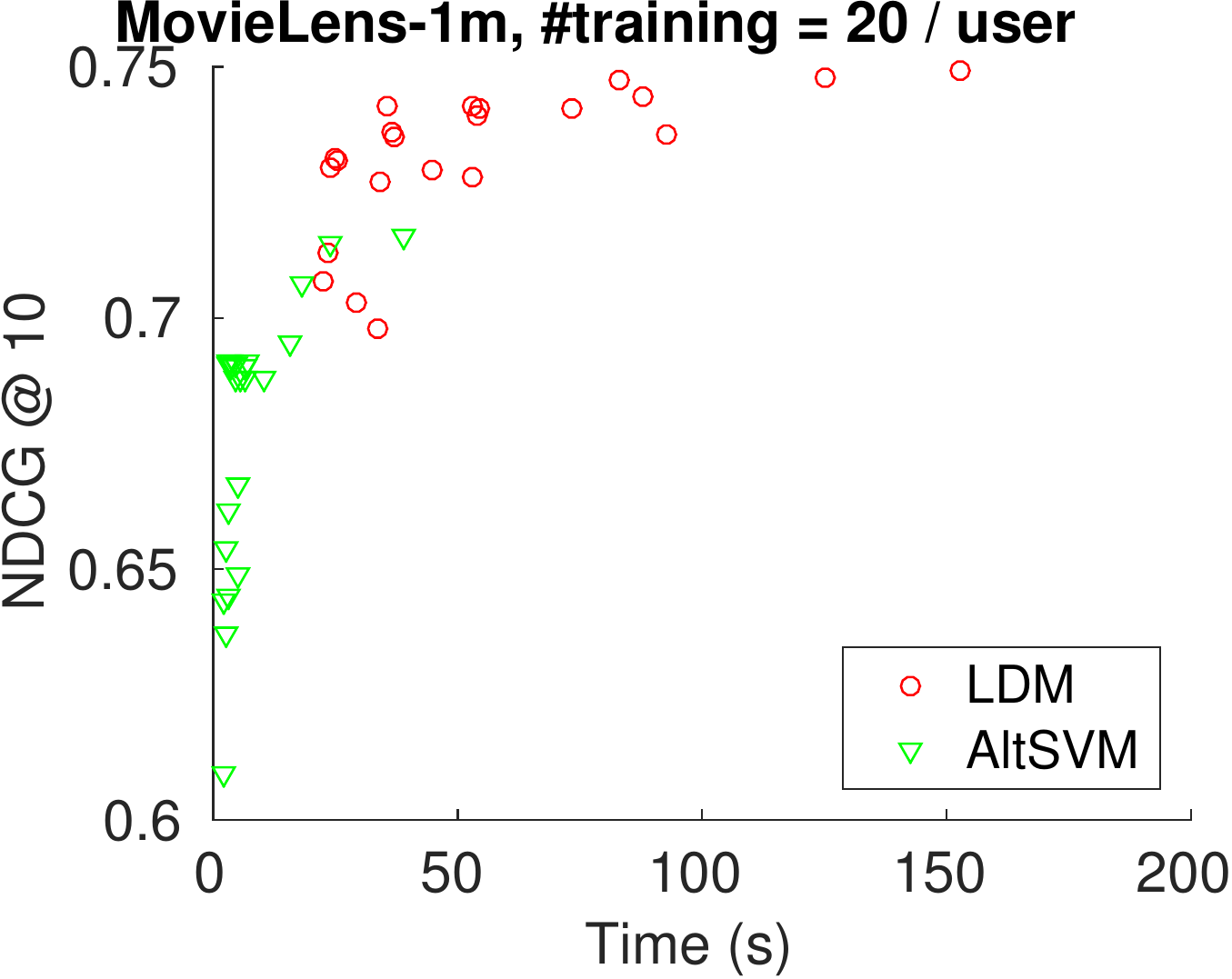}
\caption{Ranking quality NDCG@$10$ (Section \ref{sec:eval_measure}) vs. run-time under various hyperparameters for each of the methods compared. LCR and SVD are time-consuming and therefore their results with multiple sets of hyperparameters are not available on MovieLens-1m.}
\label{fig:all_param} \centering
\end{figure*}

First, we examine the influence of the {\em kd}-tree parameters on the performance of LDM, namely the number of nearest neighbors $k$ and the maximum number of distance comparisons $D$. Fig. \ref{fig:kdtree_param} shows the change in NDCG@$10$ when varying $k$ and $D$ on several small data sets (due to time constraints). In general, the ranking quality is much better with moderately large $k,D$ values than with very small $k,D$ values, but does not improve much when further increasing $k$ and $D$. Therefore, we can use sufficiently large $k,D$ values to get good ranking quality, but not too large to be computationally efficient. In the experimental comparison in Section \ref{results}, we fix the parameters to $k=64$ and $D=256$.

Next, we vary the hyperparameters in each of the four methods, and compare simultaneously the ranking quality and run-time under different hyperparameters. Ideally, a good performance of a collaborative ranking method means producing higher NDCG@$10$ scores in less time. Fig. \ref{fig:all_param} plots NDCG@$10$ against the run-time for several small data sets (due to time constraints). LDM achieves the highest NDCG@$10$ in a reasonable amount of time compared to the other methods. AltSVM is efficient but produces unsatisfactory ranking quality, which is also sensitive to its hyperparameters. For LCR, the ranking quality is acceptable but it takes considerably longer time, especially when the size of training set increases. On MovieLens-100k ($N=20$), SVD and LDM achieve similar NDCG@$10$ scores but LDM costs much shorter run-time.

\subsection{Results}
\label{results}

\begin{table*}[!t]
\caption{Benchmarking results of ranking quality NDCG@$10$ (Section \ref{sec:eval_measure}) and run-time for all the compared methods. $N$ is the number of ratings in the training set for each user.
}
\label{table:final_result} \centering
{\small
\begin{tabular}{|l||c|c|c|c||r|r|r|r|}
\hline
\multirow{2}{*}{} & \multicolumn{4}{c||}{\textbf{NDCG@$10$}} & \multicolumn{4}{c|}{\textbf{Time (seconds)}} \\
\cline{2-9} & SVD & LCR & AltSVM & LDM & SVD & LCR & AltSVM & LDM \\
\hline
MovieLens-1m, $N=10$ & 0.6836 & 0.7447 & 0.6680 & 0.7295 & 844.4 & 254.2 & 3.6 & 61.1 \\
\hline
MovieLens-1m, $N=20$ & 0.6758 & 0.7428 & 0.6879 & 0.7404 & 843.3 & 437.3 & 6.8 & 52.3 \\
\hline
MovieLens-1m, $N=50$ & 0.6178 & 0.7470 & 0.7730 & 0.7527 & 730.5 & 1168.8 & 53.0 & 37.0 \\
\hline
MovieLens-10m, $N=10$ & 0.6291 & 0.6866 & 0.6536 & 0.7077 & 24913.4 & 4544.4 & 61.2 & 1496.3 \\
\hline
MovieLens-10m, $N=20$ & 0.6201 & 0.6899 & 0.7208 & 0.7213 & 14778.5 & 6823.5 & 275.4 & 1653.1 \\
\hline
MovieLens-10m, $N=50$ & 0.5731 & 0.6830 & 0.7507 & 0.7286 & 10899.1 & 14668.5 & 648.4 & 1295.0 \\
\hline
MovieLens-10m, $N=100$ & 0.5328 & 0.7125 & 0.7719 & 0.7349 & 14648.5 & 4289.1 & 1411.2 & 832.0 \\
\hline
\end{tabular}}
\end{table*}

Now we fix the hyperparameters as described in Section \ref{algcompare}. Table \ref{table:final_result} reports the run-time and NDCG@$10$ scores for all the compared methods on the larger data sets. LDM does not achieve the highest NDCG@$10$ scores in every case, but produces robust ranking quality with decent run-time (except MovieLens-10m, $N=50$). For LCR and SVD, the time cost increases dramatically on larger data sets. AltSVM achieves superior ranking quality when the number of training ratings $N$ is large, but its performance is sensitive to the number of iterations, which in turn depends on the data set and the given tolerance parameter. We conclude that LDM is an overall competitive method that is efficient and robust to hyperparameters and the underlying data sets. Also, LDM has particular advantages when the available information is relatively few, i.e. when $N$ is small, which we consider is a more difficult problem than the cases with richer training information.

We can clearly see that the run-time of LDM increases with the number of users (due to the reliance on the user-user similarity matrix, as expected), while the run-time of AltSVM increases with the number of ratings in the training set. Further optimizing the code for LDM is an important direction in our future work to make it efficient for large-scale data sets.

We note that our method predicts all the missing ratings in the label propagation process, which is included in the timing of LDM, and therefore our methods takes a negligible amount of time for testing, especially in the real scenario where a recommendation algorithm would have to predict the ratings for all the items and return the top-rated ones to the user.

\section{Related Work}
\label{related}

Both user-based and item-based collaborative filtering \cite{userbased,itembased} can be considered as graph-based local label propagation methods. The idea of discrete harmonic extension for semi-supervised learning in general was originally proposed in \cite{zgl}. Graph-based methods were recently considered for matrix completion \cite{mcgraph2014} and top-N recommendation \cite{zhejiang} as well.
Given all the existing work, what we have presented is a continuous harmonic extension formulation for label propagation and a rigorous manifold learning algorithm for collaborative ranking.

\section{Conclusion and Discussion}
In this paper, we have proposed a novel perspective on graph-based methods for matrix completion and collaborative ranking. For each item, we view the user-user graph as a partially labeled data set (or vice versa), and our method propagates the known labels to the unlabeled graph nodes through the graph edges. The continuous harmonic extension problem associated with the above semi-supervised learning problem is defined on a user or item manifold solved by a point integral method. Our formulation can be seen as minimizing the dimension of the user or item manifold, and thus builds a smooth model for the users or items but with higher complexity than low-rank matrix approximation.
Also, our method can be fully parallelized on distributed machines, since the linear systems that need to be solved for all the items are independent with one another.
Experimental results have shown that our method has particular strength when the number of available ratings in the training set is small, which makes it promising when combined with other state-of-the-art methods like AltSVM.
Our formulation for harmonic extension in the context of matrix completion can be extended to include other constraints or regularization terms and side information as well.
An important direction is to 
further improve the efficiency of the algorithm to be comparable with recent large-scale matrix completion methods \cite{nomad,libmf}.

\label{conclusion}

 
\bibliography{LDM2}
\bibliographystyle{IEEEtranS}

\end{document}